\documentclass[runningheads]{llncs}
\usepackage{graphicx}
\usepackage{hyperref}
\usepackage{float}
\usepackage{enumitem}
\usepackage{makecell}
\usepackage{multirow}

\usepackage{bm}
\usepackage{tikz}
\usepackage{comment}
\usepackage{amsmath,amssymb} 
\usepackage{color}

\usepackage[accsupp]{axessibility}  

\usepackage[width=122mm,left=12mm,paperwidth=146mm,height=193mm,top=12mm,paperheight=217mm]{geometry}

\begin{document}

\pagestyle{headings}
\mainmatter
\def\ECCVSubNumber{1698}  

\title{Geometric Representation Learning for Document Image Rectification} 
\titlerunning{Geometric Representation Learning for Document Image Rectification}

%
\author{Hao Feng\inst{1} \and
Wengang Zhou\inst{1,2}$^{\star}$ \and  
Jiajun Deng\inst{1} \and \\
Yuechen Wang\inst{1} \and 
Houqiang Li\inst{1,2}\thanks{Corresponding Authors: Wengang Zhou and Houqiang Li.}
}
\authorrunning{H. Feng et al.}
\institute{CAS Key Laboratory of GIPAS, EEIS Department, \\University of Science and Technology of China \\
\email{\{haof,wyc9725\}@mail.ustc.edu.cn, \{zhwg,dengjj,lihq\}@ustc.edu.cn} \\
\and Institute of Artificial Intelligence, Hefei Comprehensive National Science Center
}

\maketitle

\begin{abstract}
In document image rectification, there exist rich geometric constraints between the distorted image and the ground truth one.
However, such geometric constraints are largely ignored in existing advanced solutions, which limits the rectification performance. 
To this end, we present DocGeoNet for document image rectification by introducing explicit geometric representation. 
Technically, two typical attributes of the document image are involved in the proposed geometric representation learning, \emph{i.e.}, 3D shape and textlines. 
Our motivation raises from the insight that 3D shape provides global unwarping cues for rectifying a distorted document image, while overlooking the local structure. 
On the other hand, textlines complementarily provide explicit geometric constraints for local patterns. 
The learned geometric representation effectively bridges the distorted image and the ground truth one.
Extensive experiments show the effectiveness of our framework and demonstrate the superiority of our DocGeoNet over state-of-the-art methods on both the DocUNet Benchmark dataset and our proposed DIR300 test set.
Code is available at \url{https://github.com/fh2019ustc/DocGeoNet}.
\end{abstract}


\section{Introduction}

With the popularity of smartphones, more and more people are using them to digitize document files. 
Compared to typical flatbed scanners, smartphones provides a flexible, portable, and contactless way for document image capturing. 
However, due to uncontrolled physical deformations, uneven illuminations, and various camera angles, those document images are always distorted. 
Such distortions make those images invalid in many formal review occasions, and are likely to cause the failure of the downstream applications, such as automatic text recognition, analysis, retrieval, and editing. 
To this end, over the past few years, document image rectification has become an emerging research topic. 
In this work, we focus on the geometric distortion rectification for document images, aiming to rectify arbitrarily warped documents to their original planar shape.

Traditionally, document image rectification is addressed by 3D reconstruction.
Generally, the 3D mesh of the warped document is estimated to flatten the document image.
However, such techniques are either based on auxiliary hardware~\cite{937649,brown2007,6909892,4407722} or developed with multiview images~\cite{brown2004image,4916075,yamashita2004shape,Shaodi}, which are unfriendly in personal application.
Some other methods assume a parametric model on the document surface
and optimize the model by extracting specific representations such as
shading~\cite{1561180}, boundaries~\cite{6628653}, textlines~\cite{958227,wu2002document}, or texture flow~\cite{liang2008geometric}.
However, the oversimplified parametric models 
usually lead to limited performance, and the optimization process introduces non-negligible computational cost.
Recently, deep learning based solutions~\cite{9010747,das2021end,feng2021doctr,feng2021docscanner,li2019document,liu2020geometric,8578592} have been become a promising alternative to traditional methods. 
By training a network to directly predict the warping flow, a deformed document image can be rectified by resampling the pixels in the distorted image. 
Although these methods are reported with the state-of-the-art performance, the rich geometric constraints between distorted images and ground truth ones are largely ignored.

Generally, in a document image, the texture mainly exists in textlines.
Note that there are strong geometric constraints among textlines between the distorted and ground truth image, 
that is, the curved textlines should be straight after rectification if they are horizontal textlines in a document.
In other words, textlines provide a strong cue for the rectification.
However, existing methods all just learn this prior implicitly with deep networks via the supervision on predicted warping flow, which leads to sub-optimal performance.
Besides, compared to a distorted document image, the attribute of 3D shape is a more explicit representation that directly determines the unwarping process.
The above two attributes bridge the distorted and ground truth image and complement each other:
the distribution of textlines reflects the local deformation of a document, which serves as a complement to 3D shape on local structure detail.
Based on the above motivation, we explicitly learn the geometric constraints from such attributes 
in a deep network to promote the rectification performance.


In this work, we present DocGeoNet, a new deep network for document image rectification.
DocGeoNet bridges the distorted image and its ground truth by introducing geometric constraint representation derived from document attributes.
It consists of a structure encoder, a textline extractor, and a rectification decoder.
Specifically, given a distorted document image, DocGeoNet takes the structure encoder and textline extractor to model the 3D shape of the deformed document and extract its textlines, respectively.
Then, to take advantage of the complementarity of such two attributes and leverage their direct constraints that link the distorted and ground truth image, we further fuse their representation and predict the rectification in the rectification decoder.
During the training of DocGeoNet, the learning of 3D shape, textlines, and rectification is optimized in an end-to-end way. 
Besides, considering that the 3D shape is a global attribute while the textline is a local attribute, the proposed DocGeoNet adopts a hybrid network structure, which takes advantage of self-attention mechanism~\cite{Vaswani2017AttentionIA} and convolutional operation for enhanced representation learning.
To evaluate our approach, extensive experiments are conducted on the Doc3D dataset~\cite{9010747}, DocUNet Benchmark dataset~\cite{8578592}, and our proposed challenging DIR300 Benchmark dataset.
The results demonstrate the effectiveness
of our method as well as its superiority over existing state-of-the-art methods. 

In summary, we make three-fold contributions as follows:
\begin{itemize}[noitemsep,nolistsep]
    \item 
    We present DocGeoNet, 
    a new deep network that performs explicit representation learning of the geometric constraints between the distorted and target rectified image 
    to promote the performance of document image rectification.
    \item 
    We design a new pipeline to automatically annotate the textlines of the distorted document images in training set.
    Besides,
    to reflect the effectiveness of existing works,
    we propose a new large-scale challenge benchmark dataset.
    \item
    We conduct extensive experiments to validate the merits of DocGeoNet, 
    and show state-of-the-art results on the prevalent and proposed benchmarks.
\end{itemize}

\section{Related Work}

\noindent
\textbf{Rectification Based on 3D Reconstruction.} 
Early methods first estimate the 3D mesh of the deformed document and then flatten it to a planar shape.
Brown and Seales~\cite{937649} deploy a structured light 3D acquisition system to acquire the 3D model of a deformed document.
Zhang~et al.~\cite{4407722} use a laser range scanner and perform restoration using a physical modeling technique. 
Meng~et al.~\cite{6909892} use two structured beams illuminating upon the document to recover two spatial curves of document surface. Such methods generally rely on auxiliary hardware to scan the deformed documents, which is unfriendly in daily personal use.
	
On the other hand, some methods make use of multiview images to reconstruct the 3D document model.
Tsoi~et al.~\cite{tsoi2007multi} transform the multiple views of a document to a canonical coordinate frame based on the boundaries of the document.
Koo~et al.~\cite{4916075} build the deformed surface by registering the corresponding points in two images by SIFT~\cite{lowe2004distinctive}.
Recently, You~et al.~\cite{Shaodi} propose a ridge-aware surface reconstruction method based on multiview images. 
However, in the above works, the involvement of multiview shooting limits the further applications.

Some other methods aim to reconstruct the 3D shape from a single view.
Typically, they assume a parametric model on the document surface and optimize the model by extracting specific representations, such as
    shading~\cite{1561180,wada1997shape}, boundaries~\cite{6628653}, textlines~\cite{1227630,958227,wu2002document}, or texture flow~\cite{liang2008geometric}.
Tan et al.~\cite{1561180} build the 3D shape of a book surface from the shading information. 
He et al.~\cite{6628653} extract a book boundary model to reconstruct the book surface. 
Cao et al.~\cite{1227630} and Meng et al.~\cite{meng2011metric} represent the surface as a general cylindrical surface and extract textlines to estimate the parameter of the model.

\smallskip
\noindent
\textbf{Rectification Based on Deep Learning.} 
For document image rectification, 
the first learning-based method is DocUNet~\cite{8578592}.
By training a stacked UNet~\cite{ronneberger2015u}, it directly regresses a pixel-wise displacement field to correct the geometric distortion.
Later, Li~et al.~\cite{li2019document} propose to rectify the distorted image patches first and then stitch them for rectification.
Xie~et al.~\cite{xie2020dewarping} add a smooth constraint to the learning of the pixel-wise displacement field. 
Recently, Amir~et al.~\cite{markovitz2020can} propose to learn the orientation of words in a document and Das~et al.~\cite{9010747} propose to model the 3D shape of a document with a UNet~\cite{ronneberger2015u}.
Feng~et al.~\cite{feng2021doctr} introduce transformer~\cite{Vaswani2017AttentionIA} from natural language processing tasks to improve the feature representation.
Das~et al.~\cite{das2021end} predict local deformation fields and stitch them together with global information to obtain an improved unwarping. 

Different from the above methods, in this work, we approach the document image rectification by introducing the representation learning of the geometric constraints that bridge the distorted and the rectified image, which is largely overlooked by the recent state-of-the-art methods.

\begin{figure}[t]
  \centering
  \includegraphics[width=0.95\linewidth]{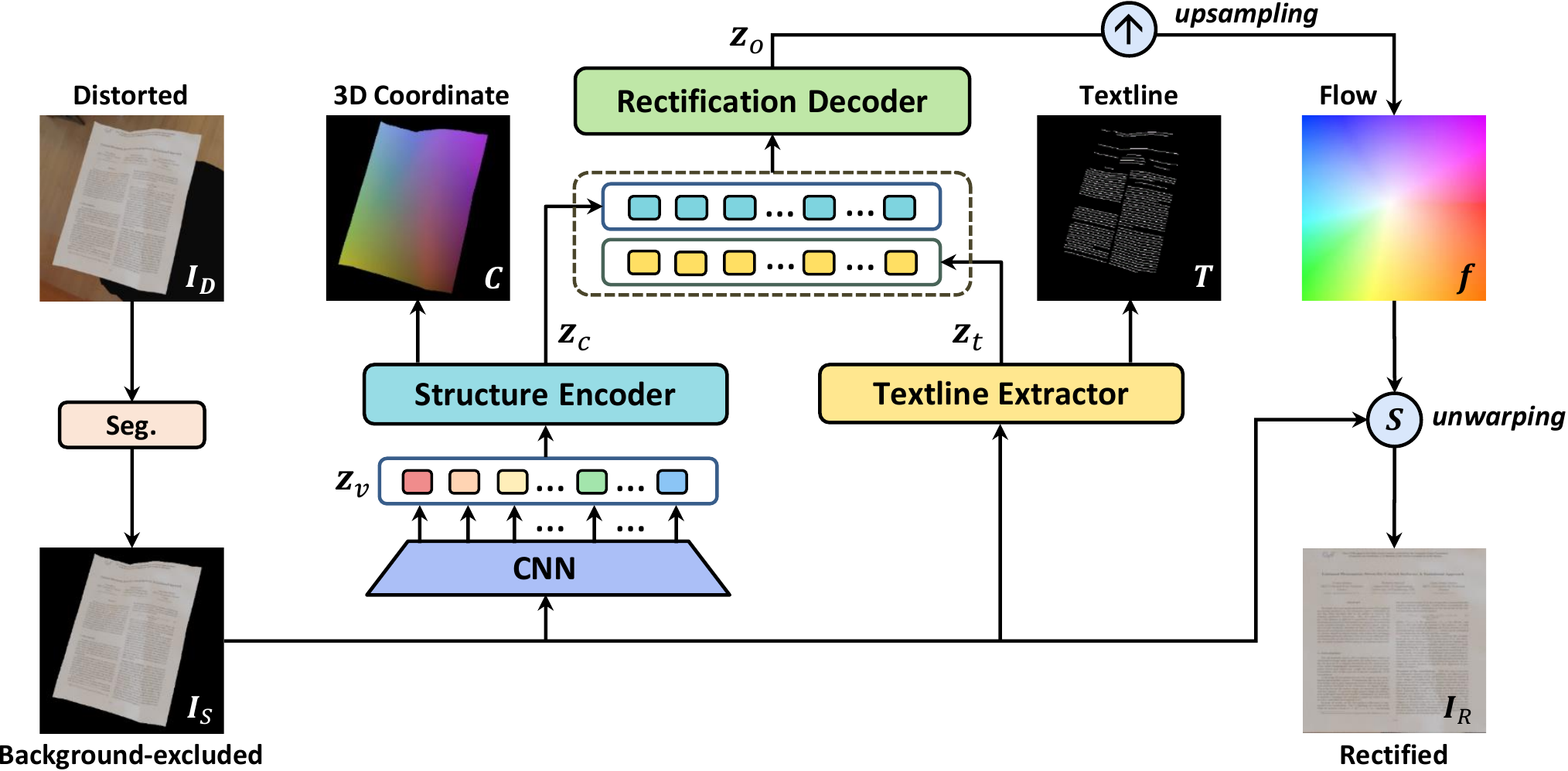}
    \caption{
    An overview of our proposed DocGeoNet. 
    It consists of three main components: 
    (1) A preprocessing module that segments the foreground document from the clustered background. 
    (2) A structure encoder and a textline extractor 
    which model the 3D shape of the deformed document and extract the curved textlines, respectively. 
    (3) A rectification decoder that estimates the warping flow for distortion rectification.}
  \label{fig:overview}
  \vspace{-0.1in}
\end{figure}

\section{Approach}
In this section, we present our Document Image Rectification Network (DocGeoNet) for geometric correction of distorted document images.
Given a distorted document image $\bm{I}_D$, 
our DocGeoNet estimates a dense displacement field $\bm{f} = (\bm{f}^x, \bm{f}^y)$ as warping flow.
Based on $\bm{f}$,
the pixel $(i, j)$ in rectified image $\bm{I}_R$ can be obtained by sampling the pixel 
$(i', j') = (i+\bm{f}^x(i), j+\bm{f}^y(j))$
in input image $\bm{I}_D$.
As shown in Fig.~\ref{fig:overview},
our framework consists of three key components:
(1) preprocessing for background removal,
(2) geometric constraint representation learning from two document attributes, 
    including 3D shape and textlines,
and (3) representation fusion and geometric rectification.
Here,
the first preprocessing stage is trained independently, 
and the latter two stages are differentiable and composed into an end-to-end trainable architecture.
In the following, we elaborate the three components separately.

\subsection{Preprocessing}
For the geometric rectification of document images,
taking the whole distorted image as input to the rectification network is a general operation.
However, it involves extra implicit learning to localize the foreground document besides predicting the rectification,
which limits the performance.
Hence, following~\cite{feng2021doctr,feng2021docscanner},
we adopt a preprocessing operation to remove the clustered background first,
thus the following network can focus on the rectification of the distortion. 

Specifically, given a distorted RGB document image $\bm{I}_D \in\mathbb{R}^{H\times W\times3}$, 
a light-weight semantic segmentation network~\cite{Qin_2020} is utilized to predict the confidence map of the foreground document.
Then, the confidence map is further binarized with a threshold $\tau$ to obtain the document region mask $\bm{M}_{\bm{I}_D} \in\mathbb{R}^{H\times W}$. 
After that, the background of $\bm{I}_D$ can be removed by element-wise matrix multiplication with broadcasting along the channels dimension of $\bm{I}_D$,
and we obtain the background-excluded document image $\bm{I}_S $. 
The preprocessing network is trained independently with a binary cross-entropy loss~\cite{de2005tutorial} as follows,
\setlength{\parskip}{0pt} 	
\begin{equation}
	\mathcal{L}_{seg} = -\sum_{i=1}^{N_p}\left[y_i\log(\hat{p_i})+(1-y_i)\log(1-\hat{p_i})\right],
\end{equation}
where $N_p=H\times W$ is the number of pixels in $\bm{I}_D$, 
$y_i$ and $\hat{p_i}$ denote the ground-truth and predicted confidence, respectively.
The obtained background-excluded document image $\bm{I}_S$ is fed into the subsequent rectification network.

\subsection{Structure Encoder and Textline Extractor}
In a document image, 
textlines are the main texture, which contain direct geometric constraints for rectification.
In other words, a distorted curved textline corresponding to a horizontal or vertical one in the ground truth should be straight after rectification.
Besides, the distribution of textlines also reflect the deformation of a document.
Therefore, textlines provide a strong cue for the rectification.
In addition, for geometric rectification,
compared to a distorted document image, the 3D shape is a more direct representation that determines the unwarping process.
Hence, we propose to model 3D shape and extract the textlines of the deformed document in the network to leverage their geometric constraints that bridge the distorted image and rectified image.

Specifically, 
as shown in Fig.~\ref{fig:overview}, 
given a background-excluded document image $\bm{I}_S$, 
we adopt two parallel sub-networks to model 3D shape and extract the textlines, respectively.
We use a transformer-based~\cite{Vaswani2017AttentionIA} sub-network for the learning of 3D shape and a CNN-based sub-network for the learning of textlines.
Such a design is adopted based on two considerations.
First, each part in a physical distorted paper is interrelated, so we introduce the self-attention mechanism~\cite{Vaswani2017AttentionIA} to capture long-distance feature dependencies.
Second, whether a pixel belongs to a textline depends more on local features, so we take advantage of convolutional operations here.
In the following, we elaborate the two sub-networks, 
\emph{i.e.}, the structure encoder and the textline extractor.

\smallskip
\noindent
\textbf{Structure Encoder.}
Given a document image $\bm{I}_S \in\mathbb{R}^{H\times W\times3}$, 
a convolutional module consisting of 6 residual blocks~\cite{he2016deep}
generates feature map $\bm{z} \in\mathbb{R}^{\frac{H}{8}\times\frac{W}{8}\times C}$,
where the channel dimension $C$ is 128.
Here the resolution of feature map decreases by $\frac{1}{2}$ every two blocks.
Then, to adapt to the sequence input form of the subsequent transformer encoder~\cite{Vaswani2017AttentionIA}, 
we flatten $\bm{z}$ into a sequence of tokens $\bm{z}_v \in \mathbb{R}^{N_v\times C}$, 
where $N_v=\frac{H}{8}\times \frac{W}{8}$ is the number of tokens.

Since that transformer layer is permutation-invariant, 
to make it sensitive to the original 2D positions of input tokens, 
we utilize sinusoidal spatial position encodings as the supplementary of visual feature.
Concretely, the position encodings are added with the query and key embedding at each transformer encoder layer.
We stack 6 transformer encoder layers and each of the encoder layers contains a multi-head self-attention module and a feed forward network.
For the $i^{th}$ encoder layer, the output representation is calculated as follows,
\begin{equation}
\begin{aligned}
    \bm{F}_0 &= [\bm{z}_v], \\
    \bm{Q}_i &= \bm{W}^Q\bm{F}_{i-1}, \bm{K}_i = \bm{W}^K\bm{F}_{i-1}, \bm{V}_i = \bm{W}^V\bm{F}_{i-1}, \\
    \bm{F}^{'}_i &= LN(MA(\bm{Q}_i,\bm{K}_i,\bm{V}_i) + \bm{F}_{i-1}), \\
    \bm{F}_i &= LN(FFN(\bm{F}^{'}_i) + \bm{F}^{'}_i),
\end{aligned}
\label{equ:transenc}
\end{equation}
where $W^Q, W^K, W^V \in \mathbb{R}^{M\times C \times C_w}$, 
$M=8$ is the number of attention heads,
$C_w=256$ denotes the feature dimension in transformer,
$MA(\cdot)$, $FFN(\cdot)$, $LN(\cdot)$ denote the multi-head attention, feed forward network, and layer normalization, respectively. 
$\bm{F}_i$ denotes the output feature of the $i^{th}$ encoder layer.
The transformer layers conducts global vision context reasoning in parallel, 
and outputs the advanced visual embedding $\bm{z}_v'$, which shares the same shape as $\bm{z}_v$.

We reshape the output feature $\bm{z}_v' \in \mathbb{R}^{N_v\times C}$ to $\frac{H}{8}\times\frac{W}{8}\times C$.
Finally, we upsample the reshaped feature map to match the resolution of the ground truth 3D coordinate map $\bm{C} \in\mathbb{R}^{H\times W\times3}$ based on bilinear sampling, 
followed by a 3 $\times$ 3 convolutional layer that reduces the channel dimension to $3$.
After that, we get the predicted 3D coordinate map $\bm{\hat{C}} \in\mathbb{R}^{H\times W\times3}$,
in which each pixel value corresponds to 3D coordinates of the document shape.

\smallskip
\noindent
\textbf{Textline Extractor.}
We segment the textlines by the per-pixel binary classification on foreground document region.
Given a background-excluded document image $\bm{I}_S \in\mathbb{R}^{H\times W\times3}$, 
a confidence map $\bm{\hat{T}} \in\mathbb{R}^{H\times W}$ with values in the range of $(0, 1)$ is predicted.
It contains the confidence of each pixel (text/non-text).

The textline extractor adopts a compact multi-scale CNN network.
It consists of a contracting part, an expansive part and a classification part.
For the contracting part,
we repeat the application of two $3\times3$ convolutional layers to encode the texture features from $\bm{I}_S$, 
each followed by a rectified linear unit (ReLU) and a $2\times2$ max pooling operation with stride 2 for downsampling.
For the expansive part,
after upsampling the feature map based on bilinear interpolation at each scale, 
we concatenate it with the corresponding feature map from the contracting path,
followed by two $3\times3$ convolutional layers and a ReLU.
In the classification part, 
a 1x1 convolutional layer followed by a Sigmoid function is used to generate the confidence map $\bm{\hat{T}} \in\mathbb{R}^{H\times W}$.

\subsection{Rectification Decoder}
\textbf{Hybrid Representation Learning.} 
To take advantage of the complementarity of the two attributes and leverage their geometric constraints that bridge the distorted and target rectified image,
we further fuse their representation and predict the rectification in the rectification decoder.
Specifically,
we first flatten the $\frac{1}{8}$ resolution representation map in expansive part of the textline extractor into a sequence of 2D features $\bm{z}_t \in \mathbb{R}^{N_v \times C_t}$. Then, we concatenate it with $\bm{z}_c \in \mathbb{R}^{N_v \times C}$, the output representation of the $4^{th}$ transformer encoder layer of the structure encoder.
The concatenated representation is feed into another 
6 transformer encoder layers to obtain the fused representation $\bm{z}_o \in \mathbb{R}^{N_v \times (C+ C_t)}$.

\smallskip
\noindent
\textbf{Rectification Estimation.}
The obtained $\bm{z}_o$ is feed into a learnable module to perform upsampling and predict high-resolution rectification estimation. 
Specifically, we first predict a coarse resolution displacement map $\bm{\hat{f}}_o\in\mathbb{R}^{(\frac{H}{8}\times\frac{W}{8})\times2}$ through a two-layer convolutional network.
Then, in analogy to~\cite{teed2020raft}, we upsample $\bm{\hat{f}}_o$ to full-resolution map $\bm{\hat{f}} \in \mathbb{R}^{H \times W \times 2}$ by taking learnable weighted combination of the $3\times 3$ grid of the coarse resolution neighbors of each pixel.

\subsection{Training Objectives}
During training,
except the preprocessing module for background removal,
the architecture of the proposed DocGeoNet is end-to-end optimized with the following objective as follows,
\begin{equation}\label{equ:loss_total}
	\mathcal{L} = \alpha \mathcal{L}_{\text{3D}} + \beta \mathcal{L}_{\text{text}} + \mathcal{L}_{\text{flow}},
\end{equation}
where $\mathcal{L}_{\text{3D}}$ denotes the regression loss on 3D coordinate map, 
$\mathcal{L}_{\text{text}}$ represents the segmentation loss of textlines,
and $\mathcal{L}_{\text{flow}}$ denotes the regression loss on warping flow.
$\alpha$ and $\beta$ are the weights associated to $\mathcal{L}_{\text{3D}}$ and $\mathcal{L}_{\text{text}}$, respectively. In the following, we present the formulation of the three loss terms. 

Specifically,
for the learning of the 3D coordinate map,
the loss $\mathcal{L}_{\text{3D}}$ is calculated as $L_1$ distance between the predicted 3D coordinate map $\hat{\bm{C}}$ and its corresponding ground truth $\bm{C}_{gt}$ as follows,
\begin{equation}
	\mathcal{L}_{3D} = \left \| \bm{C}_{gt} - \bm{\hat{C}} \right \|_1.
\end{equation}

The segmentation loss $\mathcal{L}_{\text{text}}$ for textlines is defined as a binary cross-entropy loss~\cite{de2005tutorial} as follows,
\begin{equation}
	\mathcal{L}_{text} = -\sum_{i=1}^{N_d}\left[y_i\log(\hat{p_i})+(1-y_i)\log(1-\hat{p_i})\right],
\end{equation}
where $N_d$ is the pixel number of the foreground document,
$y_i$ and $\hat{p_i}$ denote the ground-truth and predicted confidence, respectively.
Note that here we only compute the loss on the foreground document region.
One reason is that the textlines only exist in the foreground document region.
The other one is that in the input background-excluded image $\bm{I}_S$, 
the textline pixels have similar RGB values to the background,
which would confuse the network.

The loss $\mathcal{L}_{\text{flow}}$ for warping flow is defined as the $L_1$ distance between the predicted warping flow $\hat{\bm{f}}$ and its ground truth $\bm{f}_{gt}$ as follows,
\begin{equation}
	\mathcal{L}_{flow} = \left \| \bm{f}_{gt} - \hat{\bm{f}} \right \|_1.
\end{equation}

\section{DIR300 Dataset}
In this section, we present the DIR300 dataset, a new dataset for document image rectification. In the following, we first revisit the previous datasets, and then elaborate the details of the introduced one.

\subsection{Revisiting Existing Datasets}\label{sec41label}
The most widely adopted datasets in the field are Doc3D dataset~\cite{9010747} and DocUNet Benchmark dataset~\cite{8578592}. Doc3D dataset~\cite{9010747} consists of 100k synthetic distorted document images generated with the real document data and rendering software~\footnote{https://www.blender.org/}.
It is used for training the rectification model.
For each distorted document image, 
there are corresponding ground truth 3D coordinate map, depth map, and warping flow.
However, it does not contain the textline annotations, which we empirically demonstrate to be beneficial for rectification.

DocUNet Benchmark dataset~\cite{8578592} is introduced for only evaluation purpose. It contains 130 document photos captured on 65 documents, which is too small to make the evaluation results convincing. 
Besides, to the best of our knowledge, 
this is the only publicly available benchmark dataset with real image data. 
Thus, the introduction of a larger scale benchmark becomes an urgent demand.

Additionally, the $127^{th}$ and $128^{th}$ distorted
images in DocUNet Benchmark dataset~\cite{8578592} are rotated by 180 degrees, which do not match the
ground truth documents~\cite{feng2021docscanner}.
It is ignored by existing works~\cite{9010747,das2021end,feng2021doctr,jiang2022revisiting,liu2020geometric,8578592,xie2021document,xie2020dewarping,xue2022fourier}.
In our experiments, we report the results on the corrected dataset.
For clarity, we also report the performance with two mistaken samples in the supplemental material.

\begin{figure}[t]
  \centering
    \includegraphics[width=1\linewidth]{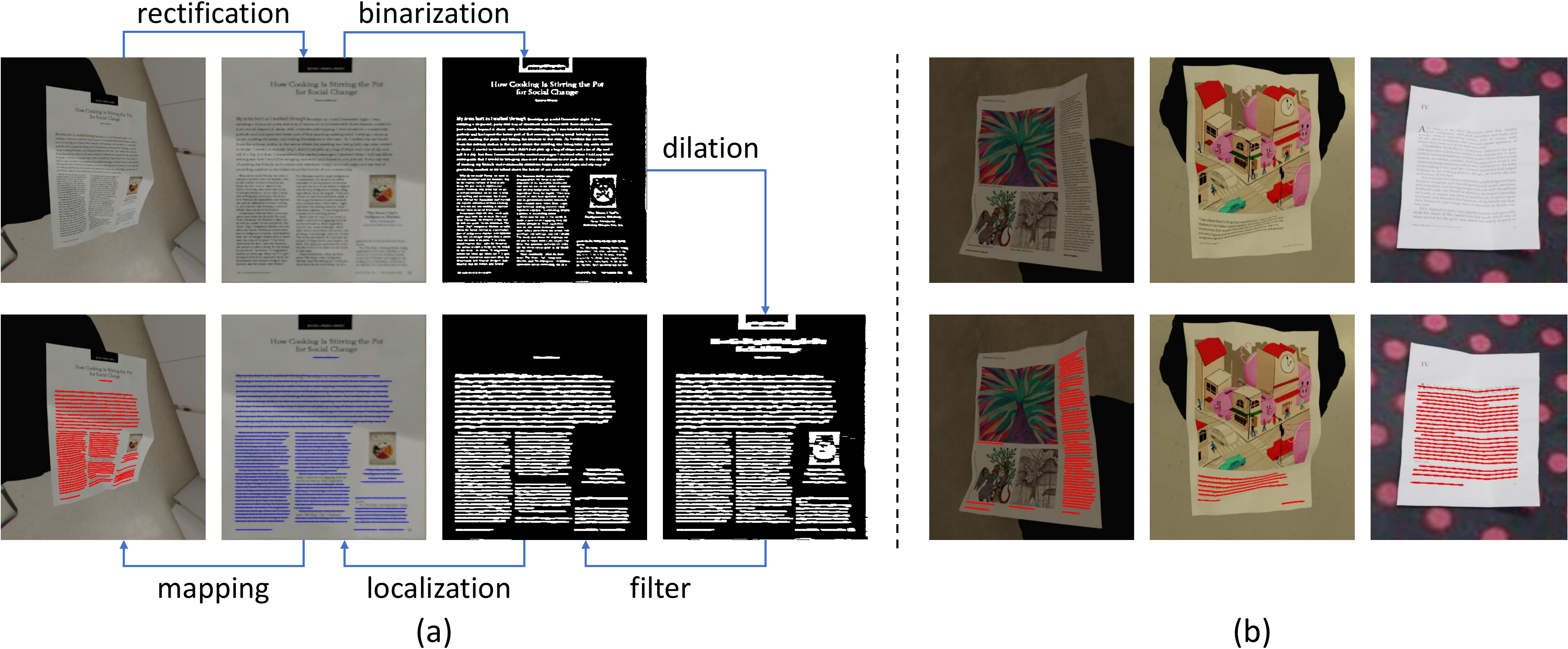}
    \vspace{-6mm}
    \caption{
    An illustration of (a) the textlines annotation process, and (b) the visualization of textline annotations in corresponding distorted document images.}
    \label{fig:textline}
    \vspace{-0.2in}
\end{figure}

\subsection{Dataset Details}
We make two-fold efforts to build the DIR300 dataset. On the one hand, we extend the synthesized Doc3D dataset~\cite{9010747} with textline annotations to build the training set. On the other hand, we capture 300 real document samples to build a larger test set against the DocUNet Benchmark dataset~\cite{8578592}.

\smallskip
\noindent\textbf{Training Set.}
Here, we describe how to generate the textline annotations on the Doc3D dataset~\cite{9010747} with fewer labour requirements. Typically, it is difficult to localize the textlines in a distorted document image, where the textlines take various shapes.
But it is easy to achieve it in a flattened document image.
Hence, we rectify all distorted images in Doc3D dataset~\cite{9010747} using the ground truth warping flow and then detect the horizontal textlines by the following steps.

Specifically, as shown in Fig.~\ref{fig:textline} (a), we first convert the rectified images to gray-scale and perform adaptive \emph{binarization} based on the local Gaussian weighted sum. 
Next, we conduct the horizontal \emph{dilation} in the binary image to get the connected regions and their corresponding bounding boxes. 
Then, we set the thresholds on the shape of bounding boxes to \emph{filter} out non-textline connected regions. 
Finally, we \emph{localize} the center and the horizontal length of the bounding boxes to generate the horizontal textlines.
After \emph{mapping} such horizontal textlines to the original distorted image using the warping flow, we obtain the curved textlines annotations. As shown in Fig.~\ref{fig:textline} (b), we visualize some textline annotation samples, where the most textlines are annotated accurately.
Notably, we note that a few annotated textlines are missed when being filtered due to their size,
but they are within the fault tolerance of the network.

\smallskip
\noindent
\textbf{Test Set.}
We build the test set of DIR300 dataset with photos captured by mobile cameras. It contains 300 real document photos from 300 documents. Compared to the DocUNet Benchmark dataset~\cite{8578592}, the distorted document images in DIR300 involve more complex background and various illumination conditions. Besides, we also increase the deformation degree of the warped documents.
The creation details are provided in the supplemental material.
To the best of our knowledge, the DIR300 test set is currently the largest real data benchmark for evaluating document image rectification.

\section{Experiments}
\subsection{Evaluation Metrics}
\noindent
\textbf{MS-SSIM.} 
The Structural SIMilarity (SSIM)~\cite{1284395} measures how similar within each patch between two images.
To balance the detail perceivability diversity that depends on the sampling density,
Multi-Scale Structural Similarity (MS-SSIM)~\cite{1292216} 
calculates the weighted summation of SSIM~\cite{1284395} across multiple scales.
Following~\cite{9010747,das2021end,feng2021doctr,jiang2022revisiting,liu2020geometric,8578592,xie2021document,xie2020dewarping}, 
all rectified and ground truth images are resized to a 598,400-pixel area.
Then, 
we build a 5-level-pyramid for MS-SSIM and the weight for each level is set as 0.0448, 0.2856, 0.3001, 0.2363, and 0.1333.

\smallskip
\noindent
\textbf{Local Distortion.} 
By computing a dense SIFT-flow~\cite{5551153}, 
Local Distortion (LD)~\cite{Shaodi} matches all the pixels from the ground truth scanned image to the rectified image.
Then, LD is calculated as the mean value of the $L_2$ distance between the matched pixels, which measures the average local deformation of the rectified image. 
For a fair comparison,
we resize all the rectified images and the ground truth images to a 598,400-pixel area, as suggested in~\cite{9010747,das2021end,feng2021doctr,jiang2022revisiting,liu2020geometric,8578592,xie2021document,xie2020dewarping}.
 
\smallskip
\noindent
\textbf{ED and CER.} 
Edit Distance (ED)~\cite{levenshtein1966binary} and Character Error Rate (CER)~\cite{morris2004and} quantify the similarity of two strings. 
ED is the minimum number of operations required to transform one string into the reference string, 
which can be efficiently computed using the dynamic programming algorithm.
The involved operations include deletions $(d)$, insertions $(i)$ and substitutions $(s)$. 
Then, Character Error Rate (CER) can be computed as follows,
\begin{equation}\label{equ:cer}
CER=(d+i+s)/{N_c} ,
\end{equation}
where $N_c$ is the character number of the reference string. 
CER represents the percentage of characters in the reference text that was incorrectly recognized in the distorted image. The lower the CER value (with 0 being a perfect score), the better the performance of the rectification quality.

We use Tesseract (v5.0.1)~\cite{4376991} as the OCR engine to recognize the text in the images.
Following DewarpNet~\cite{9010747} and DocTr~\cite{feng2021doctr},
we select 50 and 60 images
from the DocUNet Benchmark dataset~\cite{8578592}, respectively. 
Besides,
on the DIR300 test set, we select 90 images.
In such images, the text makes up the majority of content.
Since if the text is sparse in a document,
the character number ${N_c}$ (numerator) in Eq.~\eqref{equ:cer} is a small number,
leading to a large variance for CER.

\subsection{Implementation Details}
We implement the whole framework of DocGeoNet in Pytorch~\cite{paszke2017automatic}.
The preprocessing module and the following rectification module are trained independently on the extended Doc3D dataset~\cite{9010747}.
We detail their training in the following.

\setlength{\tabcolsep}{2.6pt}
\begin{table}[t]
    \scriptsize
	\caption{Quantitative comparisons of the existing learning-based methods in terms of image similarity, distortion metrics, OCR accuracy, and running efficiency on the \textbf{corrected} DocUNet Benchmark dataset~\cite{8578592}.
	``*'' denotes that the OCR metrics could not be calculated as the rectified images or models are not publicly available.
    ``$\uparrow$'' indicates the higher the better, while ``$\downarrow$'' means the opposite.}
 	\vspace{-0.05in}
	\centering
	
	\begin{tabular}{rc|cccccc}  
		Methods  & Venue & MS-SSIM $\uparrow$ &LD $\downarrow$ & ED $\downarrow$ &CER $\downarrow$  &FPS $\uparrow$
		&Para. \\  
		\Xhline{1.5\arrayrulewidth}
		Distorted & -  & 0.2459 & 20.51 & 2111.56/1552.22 & 0.5352/0.5089 & - & - \\
		DocUNet~\cite{8578592} & \emph{CVPR'18} & 0.4103 & 14.19 & 1933.66/1259.83 &0.4632/0.3966 &  0.21 & 58.6M \\
		AGUN~\cite{liu2020geometric}* & \emph{PR'18} & - & - & - & - & - & - \\
		DocProj~\cite{li2019document} & \emph{TOG'19} & 0.2946 & 18.01 & 1712.48/1165.93 & 0.4267/0.3818 & 0.11 & 47.8M \\ 
		FCN-based~\cite{xie2020dewarping} & \emph{DAS'20} & 0.4477 & 7.84 & 1792.60/1031.40 & 0.4213/0.3156 & 1.49 & 23.6M \\
		DewarpNet~\cite{9010747} &  \emph{ICCV'19} & 0.4735 & 8.39 & 885.90/525.45 & 0.2373/0.2102 & 7.14 & 86.9M \\  
		PWUNet~\cite{das2021end} &  \emph{ICCV'21} & 0.4915 & 8.64 & 1069.28/743.32 & 0.2677/0.2623 & - & - \\
		DocTr~\cite{feng2021doctr} &  \emph{MM'21} & 0.5105 & 7.76 & 724.84/464.83 &0.1832/0.1746 & 7.40 & 26.9M \\
		DDCP~\cite{xie2021document} & \emph{ICDAR'21} & 0.4729 & 8.99 & 1442.84/745.35 & 0.3633/0.2626 & \textbf{12.38} & \textbf{13.3M} \\
		FDRNet~\cite{xue2022fourier} & \emph{CVPR'22} & \textbf{0.5420} & 8.21 & 829.78/514.90 & 0.2068/0.1846 & - & - \\
		RDGR~\cite{jiang2022revisiting} &  \emph{CVPR'22} & 0.4968 & 8.51 & 729.52/420.25 & \textbf{0.1717}/0.1559 &- &- \\
		\hline
		Ours & - & 0.5040 & \textbf{7.71} & \textbf{713.94}/\textbf{379.00} & 0.1821/\textbf{0.1509}  & - & 24.8M \\ 
		
	\end{tabular}
	\label{com:b1}
\end{table}

\setlength{\tabcolsep}{4.2mm}
\begin{table}[t]
    \scriptsize
	\caption{Quantitative comparisons of the existing learning-based methods in terms of image similarity, distortion metrics, OCR accuracy on the proposed DIR300 test set.
	``$\uparrow$'' indicates the higher the better, while ``$\downarrow$'' means the opposite.}
 	\vspace{-0.05in}
	\centering
	
	\begin{tabular}{rc|cccc}  
		Methods   & Venue &MS-SSIM $\uparrow$ &LD $\downarrow$ & ED $\downarrow$ &CER $\downarrow$  \\  
		
		\Xhline{1.5\arrayrulewidth}
		
 		Distorted & - & 0.3169 & 39.58 & 1500.56 & 0.5234 \\ 
		DocProj~\cite{li2019document} & \emph{TOG'19} & 0.3246 & 30.63 & 958.89 & 0.3540 \\ 
		DewarpNet~\cite{9010747} &  \emph{ICCV'19} & 0.4921 & 13.94 & 1059.57 & 0.3557  \\  
		DocTr~\cite{feng2021doctr} &  \emph{MM'21} & 0.6160 & 7.21 & 699.63 & 0.2237 \\
		DDCP~\cite{xie2021document} & \emph{ICDAR'21} & 0.5524 & 10.95 & 2084.97 & 0.5410 \\
		\hline
		Ours & - & \textbf{0.6380} & \textbf{6.40} & \textbf{664.96} & \textbf{0.2189}  \\ 
		
	\end{tabular}
   \vspace{-0.2in}
	\label{com:b2}
\end{table}

\begin{figure}[t]
  \centering
    \includegraphics[width=1\linewidth]{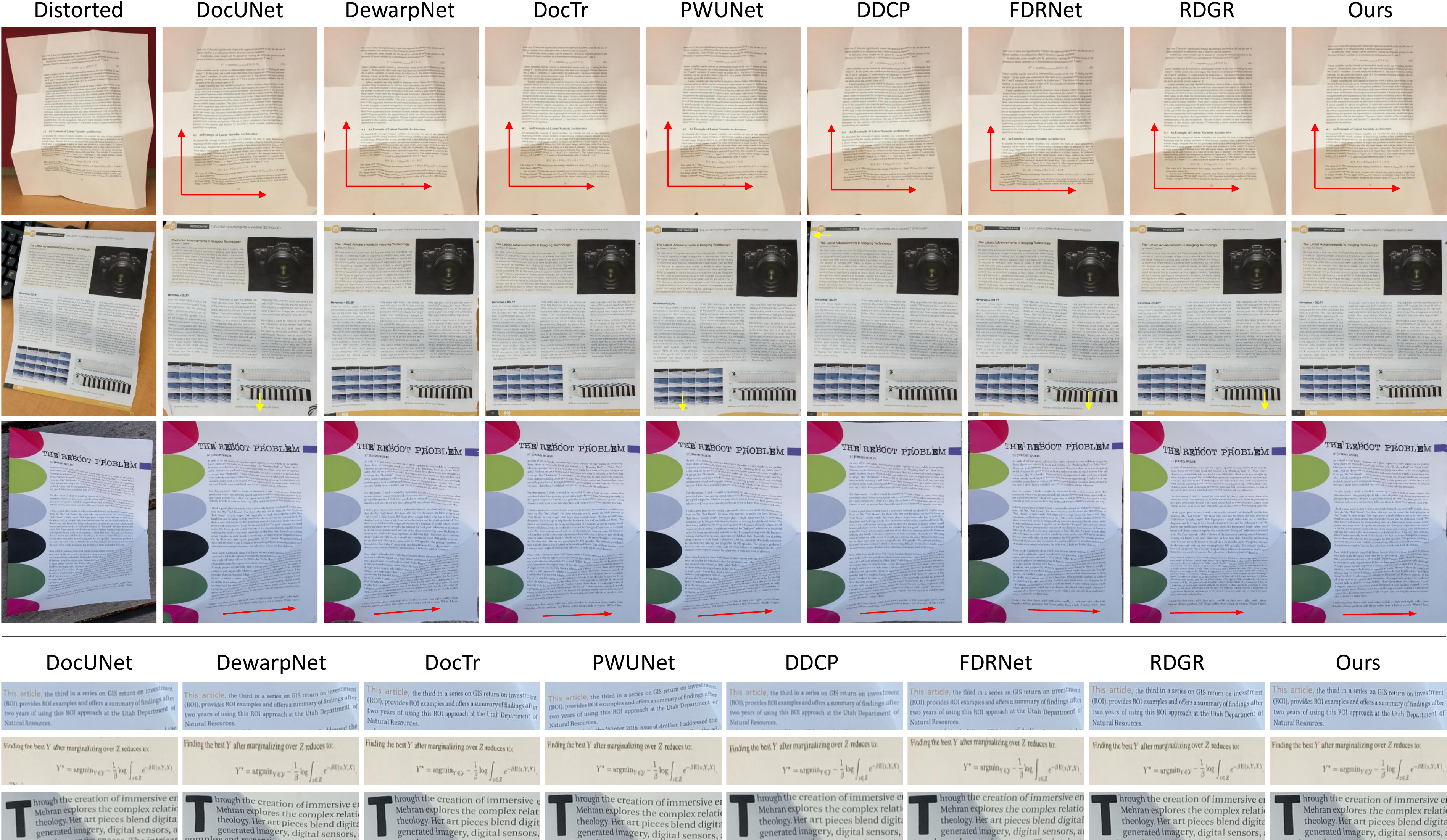}
    \vspace{-6mm}
    \caption{
    Qualitative comparisons with previous methods on the DocUNet Benchmark dataset~\cite{8578592} in terms of the rectified images and local textline detail.
    For the comparisons of the rectified images,
    we highlight the comparisons of boundary and textlines by the yellow and red arrow, respectively.}
    \label{fig:quacom1}
    \vspace{-0.06in}
\end{figure}

\begin{figure}[t]
  \centering
    \includegraphics[width=0.79\linewidth]{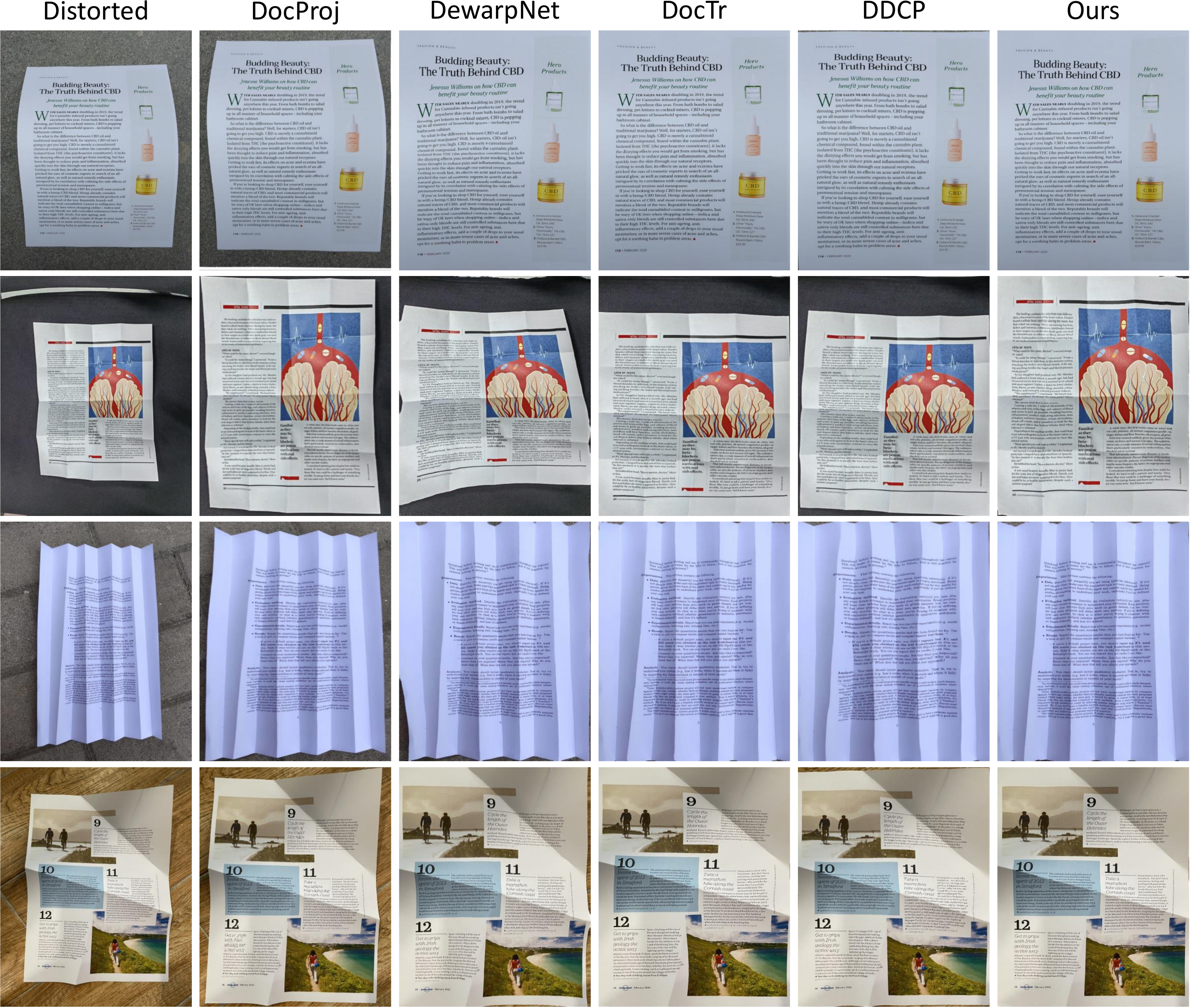}
     \vspace{-0.04in}
    \caption{
    Qualitative comparisons with previous methods on the proposed DIR300 test set in terms of the rectified images.}
    \label{fig:quacom2}
    \vspace{-0.05in}
\end{figure}

\begin{figure}[t]
   \centering
    \includegraphics[width=1\linewidth]{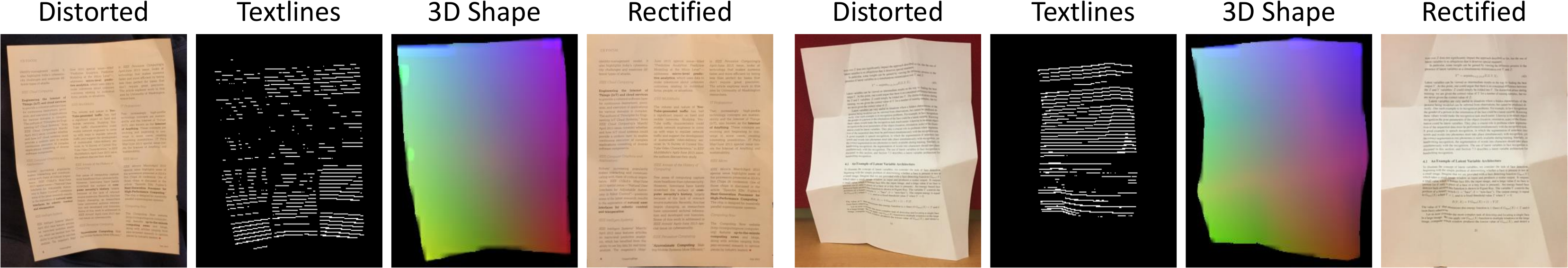}
    \vspace{-5mm}
    \caption{Visualization of two instances on the predicted textlines and 3D coordinate map by our DocGeoNet.}
    \label{fig:wc_text}
  \vspace{-0.05in}
\end{figure}

\smallskip
\noindent
\textbf{Preprocessing Module.} 
During training, 
to generalize well to real data with complex background environments, 
we randomly replace the background of the distorted document with the texture images from Describable Texture Dataset~\cite{6909856}.
We use Adam optimizer~\cite{kingma2014adam} with a batch size of 32. The initial learning rate is set as $1\times10^{-4}$, and reduced by a factor of 0.1 after 30 epochs.
The network is trained for 45 epochs
on two NVIDIA RTX 2080 Ti GPUs. 
In addition, 
the threshold $\tau$ for binarizing the confidence map is set as 0.5.

\smallskip
\noindent
\textbf{Rectification Module.} 
During training, 
we remove the background of distorted document images using the ground truth masks of the foreground document. 
To generalize well to real data with various illumination conditions, 
we add jitter in HSV color space to magnify illumination and document color variations. 
We use AdamW optimizer~\cite{loshchilov2017decoupled} with a batch size of 12 and an initial learning rate of $1\times10^{-4}$.
Our model is trained for 40 epochs on 4 NVIDIA GTX 1080 Ti GPUs. 
We set the hyperparameters $\alpha=0.2$ and $\beta=0.2$ (in Eq.~\eqref{equ:loss_total}).

\subsection{Experiment Results}
We evaluate the proposed DocGeoNet by quantitative and qualitative comparisons 
with recent state-of-the-art rectification methods.
For quantitative evaluation,
we show the comparison on distortion metrics, OCR accuracy, image similarity, and inference efficiency.
The evaluations are conducted on the corrected DocUNet Benchmark dataset~\cite{8578592} and our proposed DIR300 test set.
For clarity, in the supplementary material, we also report the performance on the DocUNet Benchmark dataset~\cite{8578592} with two mistaken samples described in Sec.~\ref{sec41label}.

\smallskip
\noindent
\textbf{Quantitative Comparisons.}
On the DocUNet Benchmark~\cite{8578592},
we compare DocGeoNet with existing learning-based methods.
As shown in Table~\ref{com:b1},
our DocGeoNet achieves a Local Distortion (LD) of 7.71 and a Character Error Rate (CER) of $15\%$,
surpassing previous state-of-the-art methods DocTr~\cite{feng2021doctr} and RDGR~\cite{jiang2022revisiting}.
In addition, 
we compare the parameter counts and inference time of processing a 1080P resolution image.
The test is conducted on an RTX 2080Ti GPU.
As shown in Table~\ref{com:b1},
the proposed DocGeoNet shows promising efficiency.

\setlength{\tabcolsep}{4.5mm}
\begin{table}[t]
    \scriptsize
	\centering
	\caption{Ablations of the architecture settings.
	SE denotes the \underline{s}tructure \underline{e}ncoder.
	TE denotes the \underline{t}extline \underline{e}xtracter.
	Here we only supervise the output warping flow.
	}
 	\vspace{-0.05in}
	\begin{tabular}{cc|cccc}
	    SE & TE & MS-SSIM $\uparrow$ & LD $\downarrow$ & ED $\downarrow$ & CER $\downarrow$  \\
       	    \Xhline{1.5\arrayrulewidth}	
	    &  &  0.4972 & 8.11 & 881.13/545.83 & 0.2231/0.1997 \\
	    $\checkmark$ & $\checkmark$ & \textbf{0.4994} & \textbf{7.99} & \textbf{781.87}/\textbf{524.90} & \textbf{0.2025}/\textbf{0.1803}  \\    
	     
	\end{tabular}
	\label{table:aba_com}
    \vspace{-0.2in}
\end{table}

On the proposed DIR300 test set,
we compare DocGeoNet with typical rectification methods with model publicly available.
As shown in Table~\ref{com:b2},
DocGeoNet outperforms the previous state-of-the-art method DocTr~\cite{feng2021doctr} 
on Local Distortion (LD) and OCR metrics,
verifying its superior rectification ability.

\smallskip
\noindent
\textbf{Qualitative Comparisons.}
The qualitative comparisons are conducted on the DocUNet Benchmark~\cite{8578592} and DIR300 test set.
To compare the local rectified detail,
we also show the comparisons of cropped local rectified text.
As shown in Fig.~\ref{fig:quacom1} and Fig.~\ref{fig:quacom2}, 
the proposed DocGeoNet shows superior rectification quality.
Specifically, for our method,
the incomplete boundaries phenomenons existing in the previous methods are to a certain extent relieved.
Besides, the rectified textlines of our method are much straighter than previous methods.
More results on the both datasets are provided in the supplementary material.

\subsection{Ablation Study}
We conduct ablation study to verify the effectiveness of the proposed DocGeoNet, 
including the architecture and the representations to learn.
The ablations are conducted on the DocUNet Benchmark dataset~\cite{8578592}.

\begin{figure}[t]
    \centering
    \includegraphics[width=0.9\linewidth]{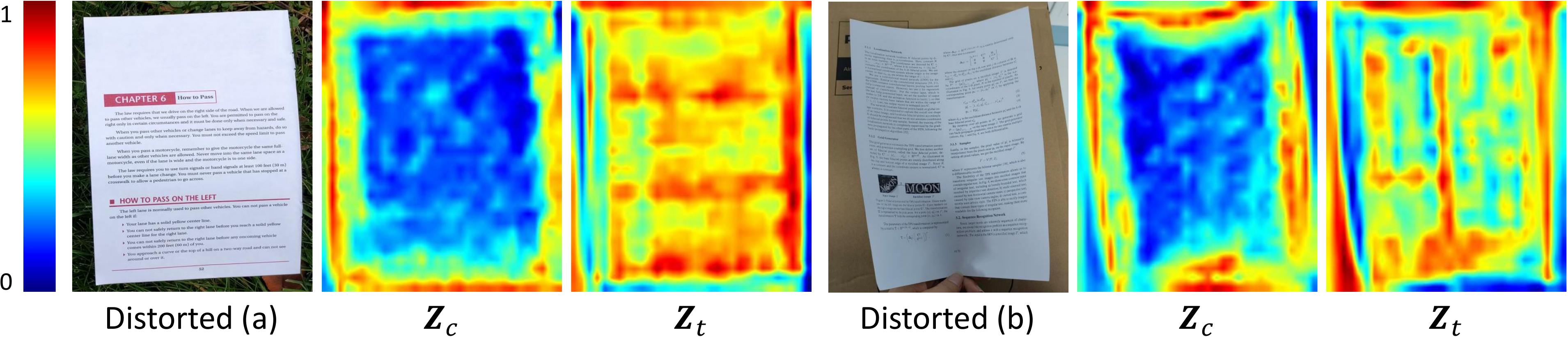}
    \vspace{-0.1in}
    \caption{Comparison of shape feature $\bm{Z}_c$ and textline feature $\bm{Z}_t$.}
  \label{fig:fea}
  \vspace{-0.05in}
\end{figure}

\smallskip
\noindent
\textbf{Architecture Setting.}
We first train a simple network without the structure encoder and textline extractor:
the background-excluded image $\bm{I}_S$ is forward to a convolutional module
and its flattened feature $\bm{z}_v$ is fed into the rectification decoder.
Then, we add the structure encoder and textline extractor while their supervisions are not deployed.
As shown in Table~\ref{table:aba_com},
the latter model obtains a slight improvement.
It is used as the baseline model for following study.

\setlength{\tabcolsep}{3mm}
\begin{table}[t]
    \scriptsize
	\centering
	\caption{Ablations of the different representation learning settings of DocGeoNet.
	}
	\begin{tabular}{cc|cccc}
	    3D Shape & Textlines & MS-SSIM $\uparrow$ & LD $\downarrow$ & ED $\downarrow$ & CER $\downarrow$  \\
        \Xhline{1.5\arrayrulewidth}
	    &  & 0.4994 & 7.99 & 784.28/524.90 & 0.2031/0.1803 \\    
	    $\checkmark$ & & \textbf{0.5067}   & 7.83 &  751.61/418.53 & 0.1843/0.1646 \\	     
	    & $\checkmark$ & 0.5053 & 7.79 & 756.75/466.58 & 0.1891/0.1693  \\	    
	    $\checkmark$ & $\checkmark$ & 0.5040 & \textbf{7.71} & \textbf{713.94}/\textbf{379.00} & \textbf{0.1821}/\textbf{0.1509} \\    
	     
	\end{tabular}
	\label{table:aba_repre}
  \vspace{-0.15in}
\end{table}

\smallskip
\noindent
\textbf{Geometric Representation.}
Based on the baseline network,
we add the supervision on the structure encoder and textline extractor, respectively.
As shown in Table~\ref{table:aba_repre},
both the representations promote the learning of rectification.
Furthermore,
the performance is better after both the supervisions are deployed.
To provide a more specific view of the predicted 3D coordinate and textlines, we showcase two examples in Fig.~\ref{fig:wc_text}.
As shown in Figure~\ref{fig:fea}, we visualize the shape feature $\bm{Z}_c$ and textline feature $\bm{Z}_t$ to help understand our primary motivation.
As we can see,
shape feature focuses more on the page boundaries and depresses the inside text content,
while textline feature does the opposite. 
The above results reveal the effectiveness of representation learning of the document attributes
that bridge the distorted image and the rectified image.

\setlength{\tabcolsep}{1.7mm}
\begin{table}[H]
    \scriptsize
	\centering
	\caption{Ablations of the different component settings of DocGeoNet.
	}
	\vspace{-0.05in}
	\begin{tabular}{c|cccc}
	    & MS-SSIM $\uparrow$  & LD $\downarrow$ & ED  $\downarrow$ & CER $\downarrow$ \\
        \Xhline{1.5\arrayrulewidth}
	    Full model & 0.5040 & \textbf{7.71} & 713.94/\textbf{379.00} & 0.1821/\textbf{0.1509}  \\
	    Preprocessing $\rightarrow$ None & 0.4843 & 8.61 & 805.63/514.60 & 0.2156/0.2003  \\    
	    Upsampling: Learnable $\rightarrow$ Bilinear  & \textbf{0.5062} & 7.77 & \textbf{706.75}/405.21 & \textbf{0.1801}/0.1523  \\    
	\end{tabular}
	\label{table:other}
	 \vspace{-0.2in}
\end{table}

\smallskip
\noindent
\textbf{Structure Modifications.}
Finally, we discuss some components of our DocGeoNet. The results are shown in Table~\ref{table:other}. 
(1) We first verify the preprocessing module that is adopted in DocGeoNet and the recent state-of-the-art method, and train a network without it. 
The results show that the performance slightly drops, which suggests that taking the whole distorted image as input burdens the network with localizing the foreground document besides the rectification prediction. 
(2) We compare the bilinear upsampling with our learnable upsampling module. 
The performance is slightly better using the learnable upsampling module. We attribute this improvement to that the coarse bilinear upsampling operation is difficult to recover the small deformations.

\section{Conclusion}
In this work, we present a novel deep network DocGeoNet for document image rectification. 
It bridges the distorted and rectified image by explicitly introducing the representation learning of the geometric constraints from two document attributes, \emph{i.e.}, 3D shape and textlines. 
Extensive experiments are conducted, and the results reveal that our DocGeoNet achieves state-of-the-art performance on the prevalent benchmark and proposed large-scale challenge benchmark. 
In the future, we will concentrate on the illumination distortion problem to further enhance the visual quality and improve the OCR accuracy.

\smallskip \noindent \textbf{Acknowledgments: } This work was supported by the National Natural Science Foundation of China under Contract 61836011 and 62021001. It was also supported by the GPU cluster built by MCC Lab of Information Science and Technology Institution, USTC.

\clearpage
%
%
\bibliographystyle{splncs04}
\bibliography{egbib}
\end{document}


\pagestyle{headings}
\mainmatter
\def\ECCVSubNumber{1698}  

\title{Geometric Representation Learning for Document Image Rectification\\
(Supplementary Material)} 
\titlerunning{Geometric Representation Learning for Document Image Rectification}
\authorrunning{H. Feng et al.}

\author{Hao Feng\inst{1} \and
Wengang Zhou\inst{1,2}$^{\star}$ \and  
Jiajun Deng\inst{1} \and \\
Yuechen Wang\inst{1} \and 
Houqiang Li\inst{1,2}\thanks{Corresponding Authors: Wengang Zhou and Houqiang Li.}
}
\authorrunning{H. Feng et al.}
\institute{CAS Key Laboratory of GIPAS, EEIS Department, \\University of Science and Technology of China \\
\email{\{fh1995,wyc9725\}@mail.ustc.edu.cn, \{zhwg,dengjj,lihq\}@ustc.edu.cn} \\
\and Institute of Artificial Intelligence, Hefei Comprehensive National Science Center
}

\maketitle

\vspace{-0.1in}
\section{Comparison with Prevalent Software}
Technically, 
the prevalent document rectification algorithms in smartphones commonly have a restriction that
the document should be a regular quadrilateral shape. 
Specifically, such techniques first detect the four corner points of the document to
localize a quadrilateral document region and then apply perspective
transformation to get the rectified image.
As a result, they can not deal with the situation when the captured document has any irregular deformations.

\begin{figure}[H]
  \vspace{-0.18in}
  \centering
    \includegraphics[width=0.95\linewidth]{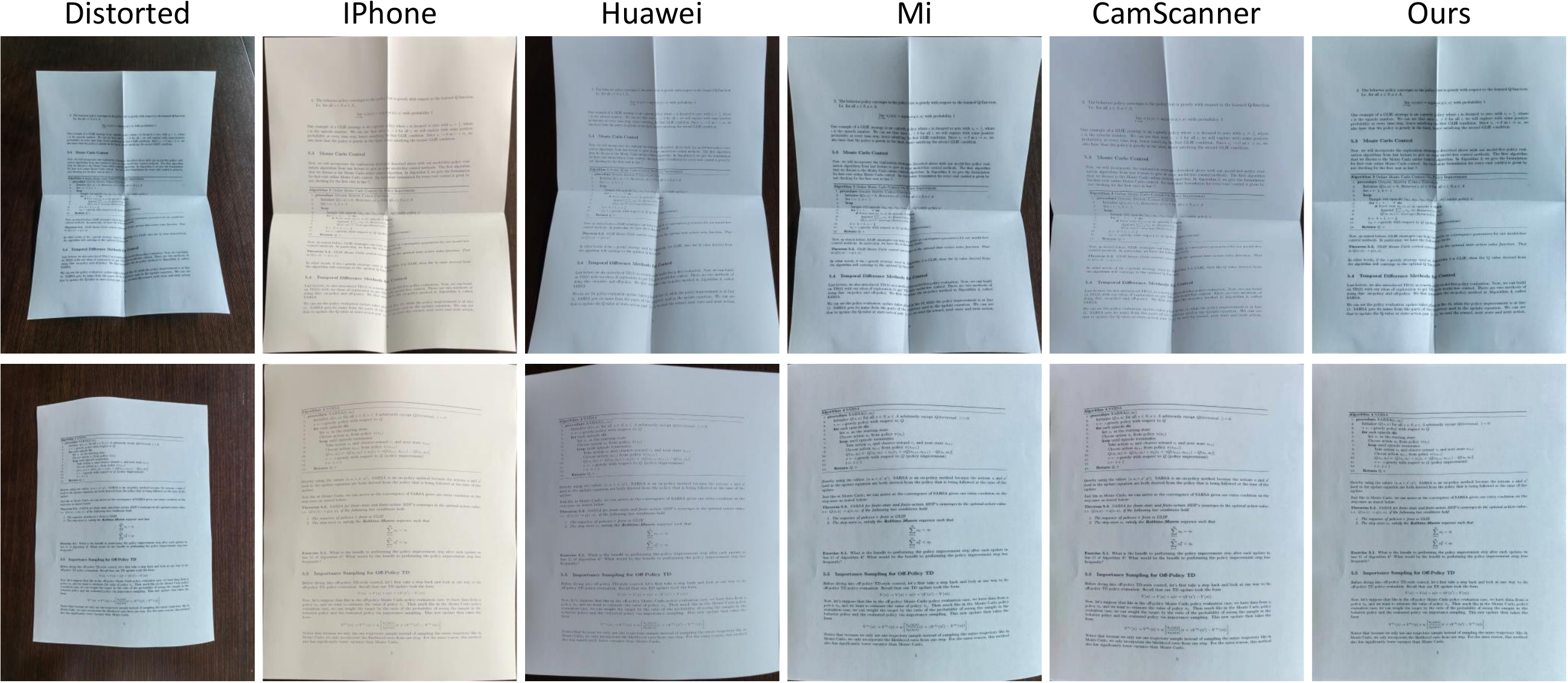} 
    \vspace{-0.1in}
    \caption{
    Qualitative comparisons of our method with the prevalent software, 
    including the CamScanner Application, the internal document rectification algorithm in smartphone of IPhone, Huawei, and Xiaomi.}
    \label{fig:smart}
\end{figure}

As shown in Figure~\ref{fig:smart},
we compare our method with the prevalent software,
including the CamScanner Application, the internal document rectification algorithm of IPhone 12, Huawei Nova 9, and Xiaomi 11.
We can see that our DocGeoNet is capable of rectifying the documents with irregular deformations.
This is because the predicted warping flow of DocGeoNet defines a non-parametric transformation, 
thus being able to represent a wide range of distortions.

\clearpage

\section{More Qualitative Results}
\begin{figure}[h]
  \vspace{-3mm}
  \centering
    \includegraphics[width=1\linewidth]{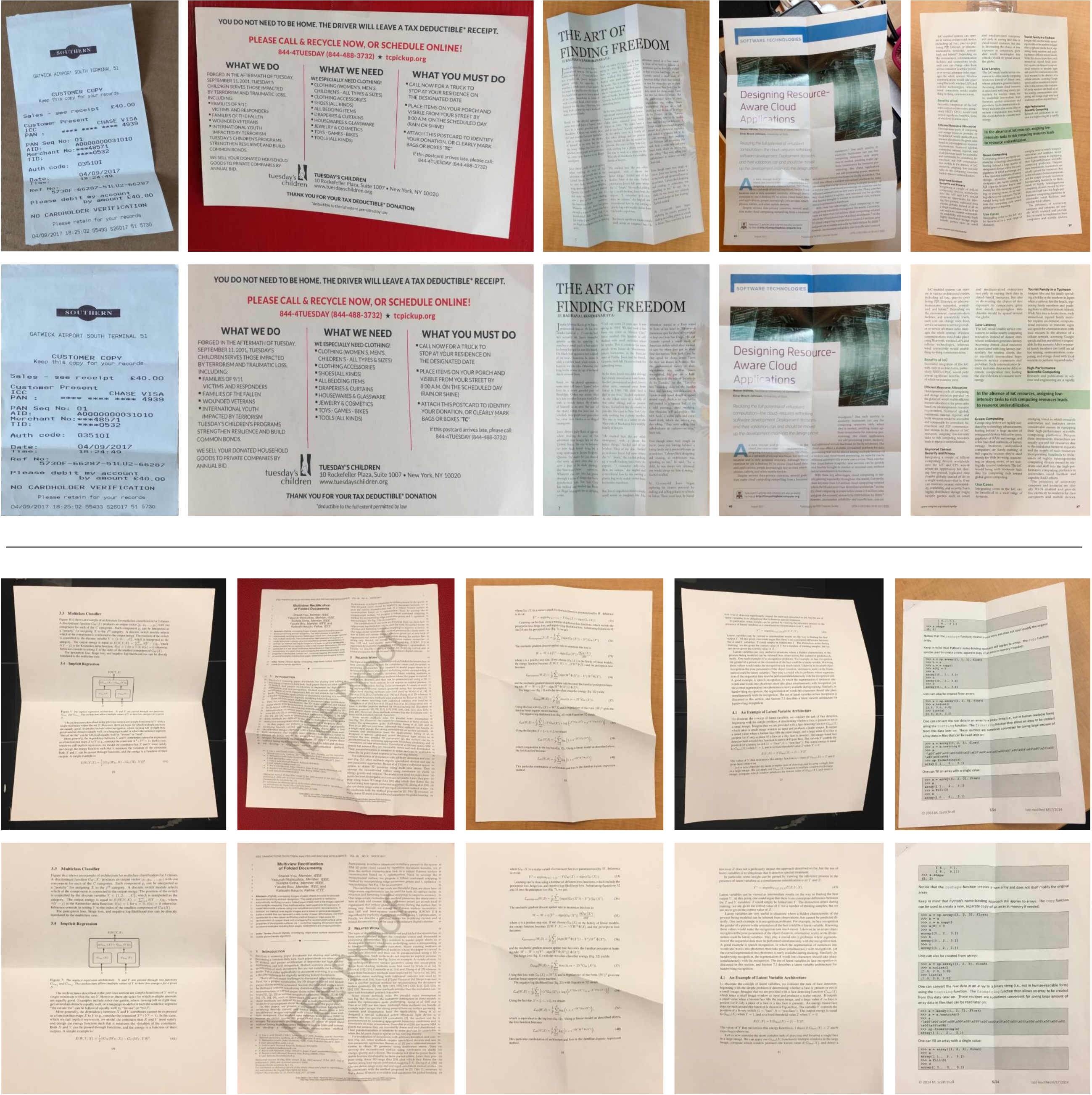} 
    \vspace{-4mm}
    \caption{
    Visualization of the rectified results on the DocUNet Benchmark dataset~[25].
    The first and third row are the input distorted images.
    The second and bottom row show their corresponding rectified results.}
    \label{fig:DOCUNET}
\end{figure}

As shown in Figure~\ref{fig:DOCUNET}, 
we present more qualitative rectified results on the DocUNet Benchmark dataset~[25].
Besides, as shown in Figure~\ref{fig:DIRNet},
we provide more rectified results on real distorted document photos.
As we can see, the proposed DocGeoNet shows superior rectification quality.

Particularly, as shown in Figure~\ref{fig:DIRNet},
the distorted images show various physical deformations, backgrounds, and illumination conditions.
These photos are captured under various indoor (during day and night) and outdoor scenes.
Besides, the used documents contain text, tables, figures, or their mixture.

\begin{figure}[H]
  \centering
    \includegraphics[width=1\linewidth]{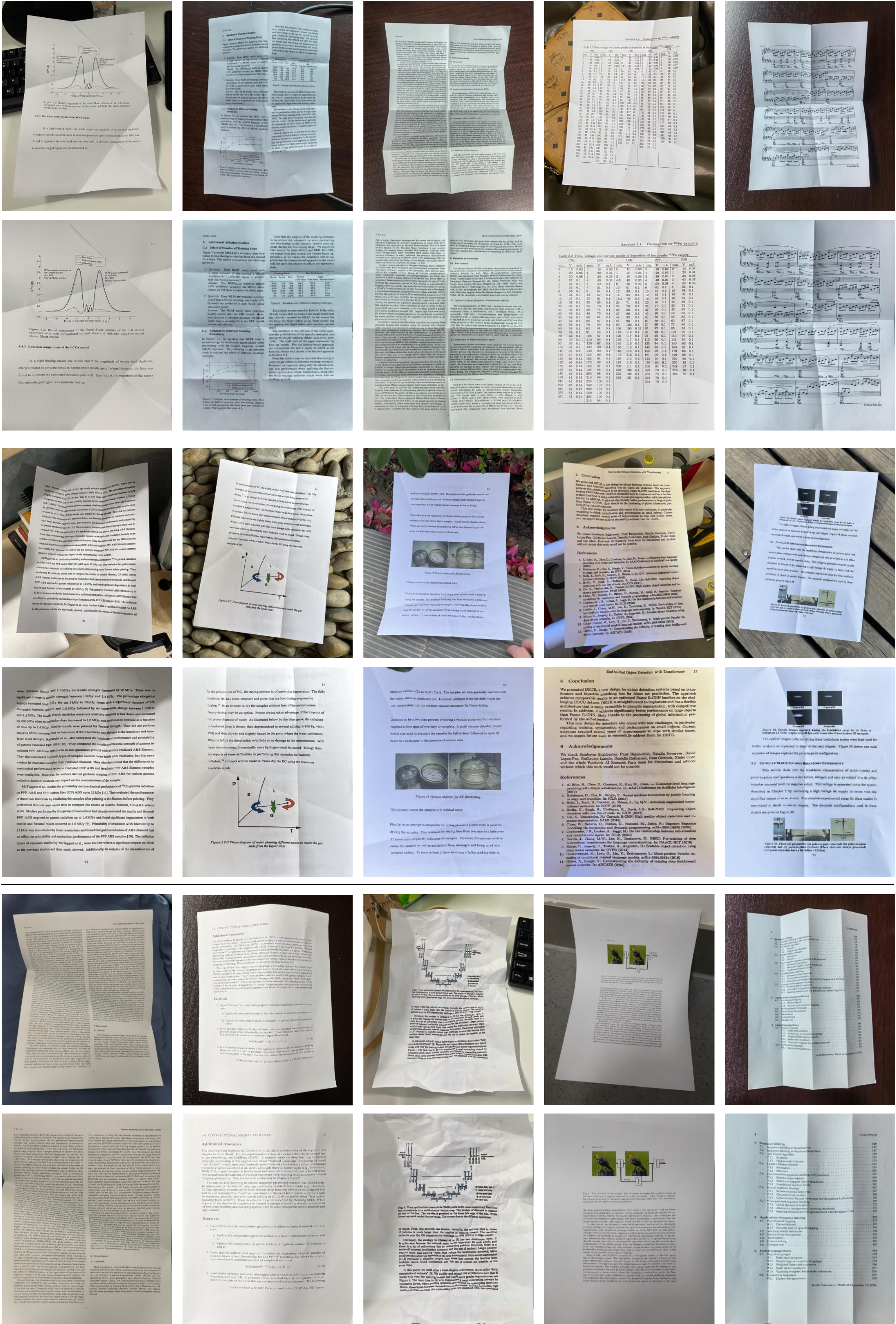} 
    \vspace{-4mm}
    \caption{
    Visualization of the rectified results on real distorted document photos under various conditions.
    The first, third and fifth row are the distorted images, and
    the remaining rows are their rectified results.}
    \label{fig:DIRNet}
\end{figure}

\section{DIR300 Dataset}
Furthermore, we present the detail about the creation of the DIR300 test set.
Concretely,
the images are firstly taken by three people with three different cellphones,
including iPhone 12, Huawei Nova 9, and Mi 11.
Each person captures 100 photos.
Secondly,
to involve various backgrounds and illumination conditions, 
for each person,
the indoor and outdoor scenes account for $70\%$ and $30\%$, respectively.
For indoor scenes, $20\%$ are taken in the evening.
Thirdly, in terms of distortions, $40\%$ samples involve random curving;
$40\%$ samples contain random folds;
$10\%$ samples are flat; the remaining $10\%$ are heavily crumpled.
Here we do not fix the scenes for a certain distortion.

Note that the ground truth images are captured before the collection of the distorted images.
Specifically, we put the regular rectangular document on a plane.
Then, we align the four corner points of the rectangular document and then get a perfect rectification.
The rectified image is taken as the ground truth.
Another way is adopted by the successful DocUNet Benchmark dataset~[25] which scanned the printed document to image as GT,
but a perfect alignment is still difficult due to the scanning error.

As shown in Figure~\ref{fig:DIRNet}, we present more qualitative rectified results on the DIR300 test set.
It can be seen that, the proposed DocGeoNet shows superior rectification performance.

\section{OCR Visualizations}
\begin{figure}[H]
  \vspace{-3.5mm}
  \centering
    \includegraphics[width=0.75\linewidth]{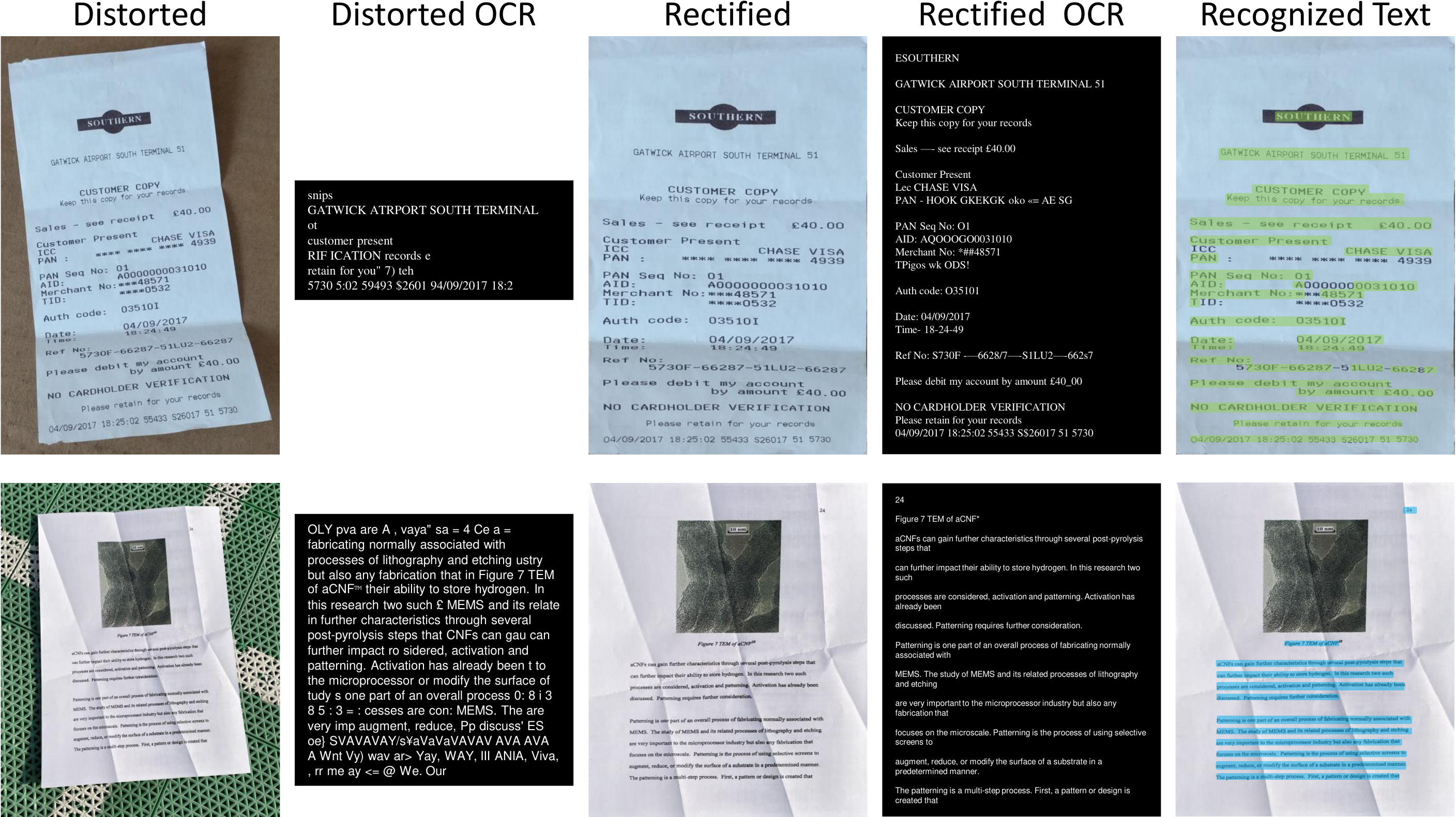} 
    \vspace{-2mm}
    \caption{
    Visualization of two instances about the impact of the geometric rectification on the OCR performance.
    The second and fourth column show the recognized text of the distorted image and the rectified image of the proposed DocGeoNet, respectively.
    Besides, we highlight the correct recognized text of DocGeoNet in the fifth column.
    }
    \label{fig:ocr}
\end{figure}

To reveal the impact of the geometric rectification on the OCR performance, 
we further visualize their OCR results, respectively.
We use the Tesseract~[33] as the OCR engine.
As shown in Figure~\ref{fig:ocr},
after the geometric rectification,
the OCR performance makes remarkable improvements.

\begin{figure}[H]
  \centering
    \includegraphics[width=0.95\linewidth]{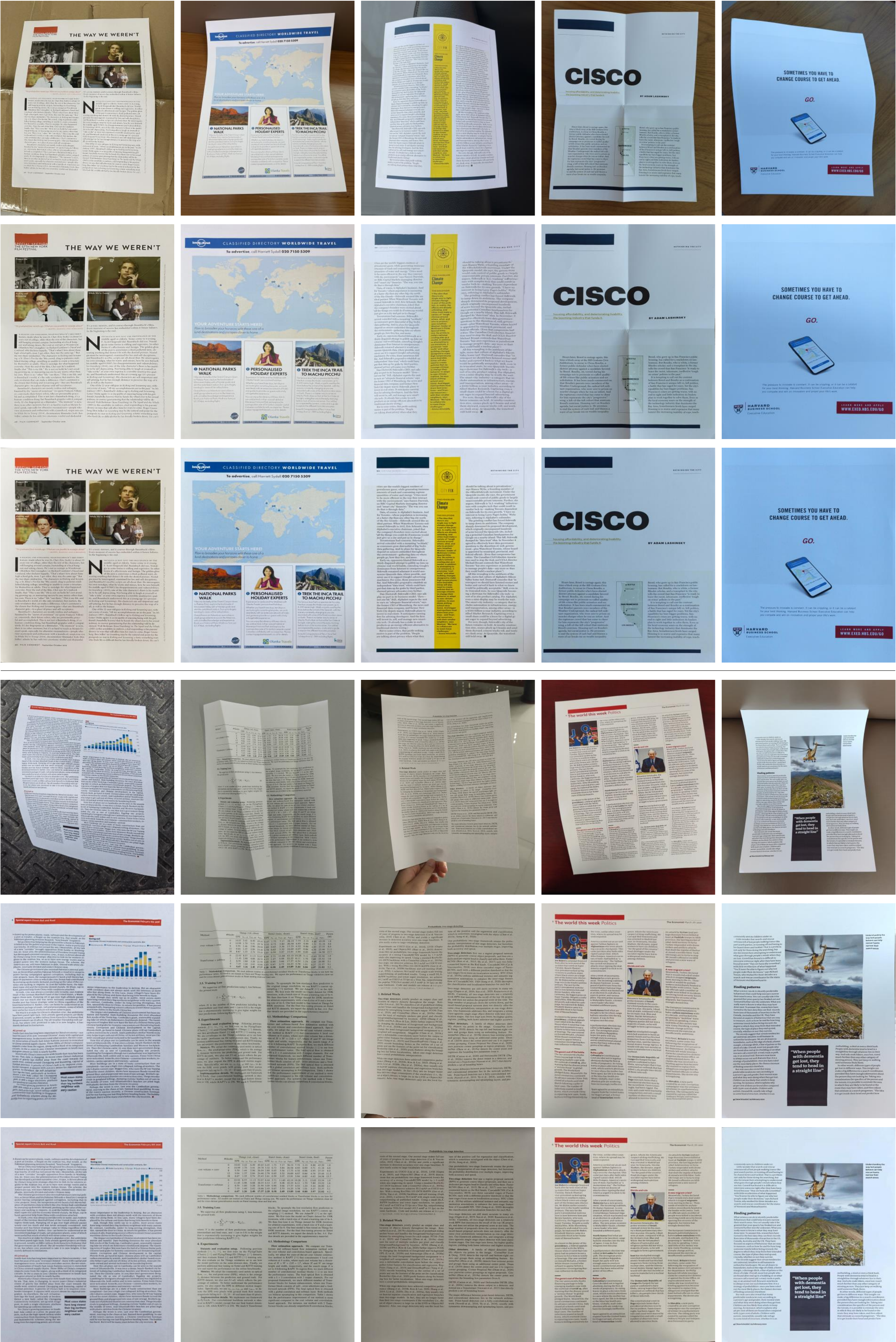} 
    \vspace{-1mm}
    \caption{
    Visualization of the rectified results on the DIR300 test set.
    The first and fourth row are the distorted images.
    The second and fifth row are their rectified results.
    The remaining are their corresponding ground truth.}
    \label{fig:DIRNet}
\end{figure}

\section{Efficiency Analysis}
In this section, we discuss the efficiency of the proposed DocGeoNet and other learning-based methods.
As shown in Table~1 in the manuscript, 
DocGeoNet is efficient in terms of inference time and parameter count compared with existing learning-based methods.
One of reasons is that we predict the backward warping flow directly that is used to sample the pixels from the input distorted image for rectification,
following~[6,9,10]. 
In contrast, previous work DocUNet~[25] predicts the forward warping flow instead, 
which has to be converted to the backward warping flow first using the unstructured points. 
In addition, 
DocProj~[19] crops the distorted image into patches first and then rectifies the patches to perform rectification. 
However, the rectification of input distorted patches and the stitching of backward warping flow patches heavily increase the computational cost.

\section{Performance on DocUNet Benchmark}
In Table 1 of the manuscript, we report the performance on the corrected DocUNet Benchmark dataset~[25].
In this section, for clarity, we also report the results on the DocUNet Benchmark dataset [25] with two mistaken image samples.
The results are shown in Table~\ref{com:b1M}.
Note that the two mistaken images are not contained in the sub-set for the OCR evaluation.
Hence, they only affect the evaluation of MS-SSIM and LD.

\setlength{\tabcolsep}{2.6pt}
\begin{table}[H]
    \scriptsize
	\caption{Quantitative comparisons of the existing learning-based methods in terms of image similarity, distortion metrics, OCR performance, and running efficiency on the DocUNet Benchmark dataset~[25] \textbf{with two mistaken image samples}.
    ``$\uparrow$'' indicates the higher the better, while ``$\downarrow$'' means the opposite.}
 	\vspace{-0.05in}
	\centering
	
	\begin{tabular}{rc|cccccc}  
		Methods  & Venue & MS-SSIM $\uparrow$ &LD $\downarrow$ & ED $\downarrow$ &CER $\downarrow$  &FPS $\uparrow$
		&Para. \\  
		\Xhline{1.5\arrayrulewidth}
		Distorted & -  & 0.2464 & 20.51 & 2111.56/1552.22 & 0.5352/0.5089 & - & - \\
		DocUNet~[25] & \emph{CVPR'18} & 0.4094 & 14.22 & 1933.66/1259.83 &0.4632/0.3966 &  0.21 & 58.6M \\
		AGUN~[22] & \emph{PR'18} & 0.4491 & 12.06 & - & - & - & - \\
		DocProj~[19] & \emph{TOG'19} & 0.2928 & 18.19 & 1712.48/1165.93 & 0.4267/0.3818 & 0.11 & 47.8M \\ 
		FCN-based~[42] & \emph{DAS'20} & 0.4361 & 8.50 & 1792.60/1031.40 & 0.4213/0.3156 & 1.49 & 23.6M \\
		DewarpNet~[6] &  \emph{ICCV'19} & 0.4692 & 8.98 & 885.90/525.45 & 0.2373/0.2102 & 7.14 & 86.9M \\  
		PWUNet~[7] &  \emph{ICCV'21} & 0.4879 & 9.23 & 1069.28/743.32 & 0.2677/0.2623 & - & - \\
		DocTr~[9] &  \emph{MM'21} & 0.5085 & 8.38 & 724.84/464.83 &0.1832/0.1746 & 7.40 & 26.9M \\
		DDCP~[41] & \emph{ICDAR'21} & 0.4706 & 9.51 & 1442.84/745.35 & 0.3633/0.2626 & \textbf{12.38} & \textbf{13.3M} \\
		FDRNet~[43] & \emph{CVPR'22} & \textbf{0.5440} & 8.75 & 829.78/514.90 & 0.2068/0.1846 & - & - \\
		RDGR~[14] &  \emph{CVPR'22} & 0.4929 & 9.11 & 729.52/420.25 & \textbf{0.1717}/0.1559 &- &- \\
		\hline
		Ours & - & 0.5027 & \textbf{8.37} & \textbf{713.94}/\textbf{379.00} & 0.1821/\textbf{0.1509}  & - & 24.8M \\ 

	\end{tabular}
	\label{com:b1M}
\end{table}

\section{Metric Analysis}
During the experiments, we find that the SSIM~[38] is not a very appropriate metric for document image rectification.
Document image rectification is not a pixel-aligned task, different them the typical pixel-aligned tasks, such as deraining and denoising.
SSIM~[38] and MS-SSIM~[39] are designed to capture the perceptual distortion of images with respect to a reference image.
Blur, noise, color shifts, and halos are the types of artifacts they are designed to capture, rather than geometric distortion (i.e., misalignment of the pixels between a reference and a corrupted image).
For future works, we recommend removing it and leaving the other metrics. 
If the authors think they should keep it to be consistent with previous work, we recommend adding a note that that is the purpose of providing that number.
A typical example is FDRNet~[43]. The SSIM score in Table 1 in the manuscript is smaller than that in Table 1 in this supplementary material. However, Table 1 in the manuscript shows the performance on the corrected DocUNet Benchmark dataset~[25].

\section{Limitation Discussion}
Existing methods all limit the size of background area of the distorted images when training the rectification networks.
This is because the background area of distorted images in DocUNet Benchmark dataset~[25] is small.
As a result, when increasing the camera distance, the background area becomes larger and the performance drops.
It is the same with our method.
We hope future works can propose more robust methods.
